\documentclass[sigconf]{acmart}
\usepackage{xspace}
\usepackage[normalem]{ulem}

\newcommand{\ccreid}{CC-ReID\xspace}
\newcommand{\reid}{Re-ID\xspace}

\newcommand\etal{\textit{et al.}\xspace}

\newcommand\modelnameshort{SCNet\xspace}
\newcommand\ie{i.e.\xspace}

\AtBeginDocument{%
  }

\copyrightyear{2023}
\acmYear{2023}
\setcopyright{acmlicensed}

\acmConference[MM '23] {Proceedings of the 31st ACM International Conference on Multimedia}{October 29--November 3, 2023}{Ottawa, ON, Canada.}
\acmBooktitle{Proceedings of the 31st ACM International Conference on Multimedia (MM '23), October 29--November 3, 2023, Ottawa, ON, Canada}
\acmPrice{15.00}
\acmISBN{979-8-4007-0108-5/23/10}
\acmDOI{10.1145/3581783.3612416}
\usepackage{multirow}
\usepackage{comment}
\usepackage{balance}
\usepackage{hyperref}
\hypersetup{colorlinks=true,
urlcolor=pink,
filecolor=pink,
citecolor=pink, 
linkcolor=pink}

\begin{document}
\begin{sloppypar}

\title{Semantic-aware Consistency Network for Cloth-changing \\ Person Re-Identification}
\renewcommand{\shorttitle}{Semantic-aware Consistency Network for Cloth-changing Person Re-Identification}

\author{Peini Guo}
\affiliation{
\institution{Shenzhen Graduate School}
\country{Peking University, China}
}
\email{guopeini@stu.pku.edu.cn}

\author{Hong Liu}
\authornote{Corresponding Author.}
\affiliation{
\institution{Shenzhen Graduate School}
\country{Peking University, China}
}
\email{hongliu@pku.edu.cn}

\author{Jianbing Wu}
\affiliation{
\institution{Shenzhen Graduate School}
\country{Peking University, China}
}
\email{kimbing.ng@stu.pku.edu.cn}

\author{Guoquan Wang}
\affiliation{
\institution{Shenzhen Graduate School}
\country{Peking University, China}
}
\email{guoquanwang@stu.pku.edu.cn}

\author{Tao Wang}
\affiliation{
\institution{Shenzhen Graduate School}
\country{Peking University, China}
}
\email{taowang@stu.pku.edu.cn}

\renewcommand{\shortauthors}{Peini Guo et al.}

\begin{abstract}
Cloth-changing Person Re-Identification (\ccreid) is a challenging task that aims to retrieve the target person across multiple surveillance cameras when clothing changes might happen.
Despite recent progress in \ccreid, existing approaches are still hindered by the interference of clothing variations since they lack effective constraints to keep the model consistently focused on clothing-irrelevant regions.
To address this issue, we present a Semantic-aware Consistency Network (\modelnameshort) to learn identity-related semantic features by proposing effective consistency constraints.
Specifically, we generate the black-clothing image by erasing pixels in the clothing area, which explicitly mitigates the interference from clothing variations.
In addition, to fully exploit the fine-grained identity information, a head-enhanced attention module is introduced, which learns soft attention maps by utilizing the proposed part-based matching loss to highlight head information.
We further design a semantic consistency loss to facilitate the learning of high-level identity-related semantic features, forcing the model to focus on semantically consistent cloth-irrelevant regions.
By using the consistency constraint, our model does not require any extra auxiliary segmentation module to generate the black-clothing image or locate the head region during the inference stage.
Extensive experiments on four cloth-changing person \reid datasets (LTCC, PRCC, Vc-Clothes, and DeepChange) demonstrate that our proposed \modelnameshort makes significant improvements over prior state-of-the-art approaches.
Our code is available at: \href{https://github.com/Gpn-star/SCNet}{https://github.com/Gpn-star/SCNet}.
\end{abstract}

\begin{comment}
The raw RGB image and black-clothing image are utilized to extract color-based appearance features and cloth-irrelevant features, respectively.
\end{comment}





\maketitle
\section{Introduction} \label{sec:introduction}
\begin{figure}[t]
  \includegraphics[width=\linewidth,height=2in]{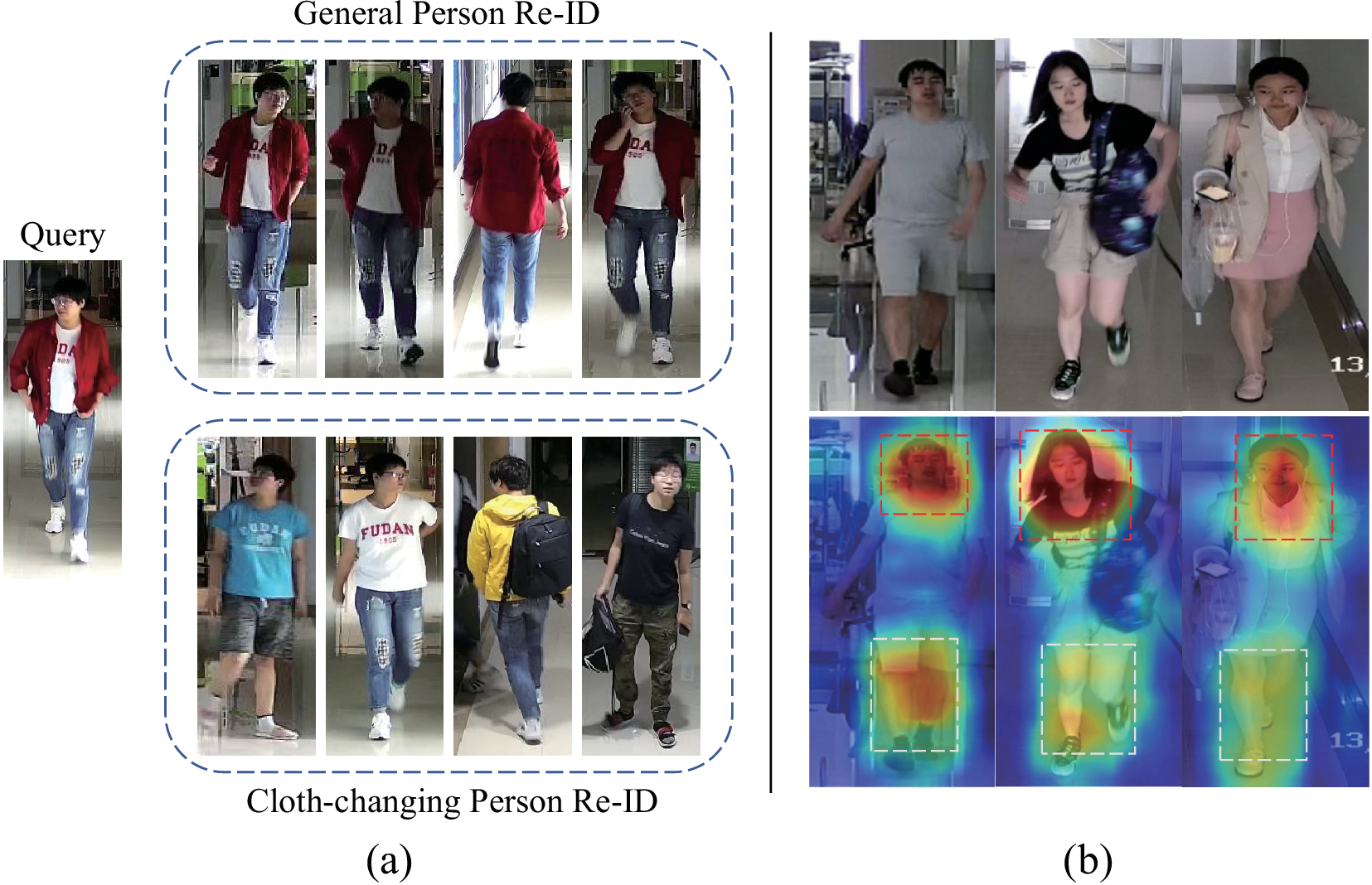}
  \caption{(a) Difference between the general person \reid and cloth-changing person \reid tasks. (b) Some examples of the semantic consistency of features.} 
  \label{fig:intro}
\end{figure}

Person Re-Identification (\reid) aims to retrieve the same target person across different viewpoints and cameras, which plays an important role in intelligent surveillance, suspect tracking, and human-computer interaction ~\cite{yeDeepLearningPerson2022}.
In recent years, general person \reid methods \cite{sun2018beyond, he2021transreid, hou2019interaction} have achieved significant performance improvements, which mainly focus on learning discriminative appearance features, tackling viewpoint variations and occlusions.
However, as shown in Fig. \ref{fig:intro} (a), most of these approaches are based on the assumption that the same pedestrian exhibits unchanged clothes and accessories at different times and places, so they rely heavily on the appearance of pedestrians.
In real-life scenarios, there is a high probability that people will change their clothing after a long time interval, especially criminal suspects who may deliberately modify their appearance to avoid being tracked.
Consequently, the performance of these methods in practical scenarios can suffer significant degradation.

To mitigate the impact of clothing changes on the person \reid, existing cloth-changing person \reid (\ccreid) methods often focus on exploiting discriminative biological cues, such as silhouette sketches \cite{liu2022long, yang2019person}, body shape \cite{hong2021fine, qian2020long}, face \cite{wan2020person, shi2022iranet}, and gait information \cite{jin2022cloth}.
Hong \etal \cite{hong2021fine} propose a dual-stream framework that enables mutual learning of shape and appearance to complement clothing-agnostic knowledge in appearance features. 
Nevertheless, the body shape extracted with the human parsing network is often affected by the size of the clothes, which inevitably brings additional interference.
M2Net \cite{liu2022long} is proposed for simultaneously employing human parsing, contour, and appearance information to learn features robust to clothing variations.
However, it simply concatenates multi-modality features in the channel dimension, introducing redundant information and increasing computational complexity.
In \cite{gu2022clothes}, the authors propose CAL, a clothes-based adversarial loss that forces the backbone of the \reid model to learn clothes-irrelevant features by penalizing its predictive power for clothes.
But this method requires the collection of clothing labels, which is a time-consuming process in practice.
In addition, some methods based on metric learning \cite{gu2022clothes, li2022cloth} and data augmentation \cite{jia2022complementary, shu2021semantic} have been proposed to address the \ccreid problem. 
Although these approaches mitigate the effect of clothing changes to some extent, they still lack effective constraints to ensure that the model consistently focuses on cloth-irrelevant regions.
Moreover, they do not adequately consider the semantic consistency of features, which is critical for accurately identifying pedestrians under clothing variations.

Semantic consistency refers to the robustness of the semantic information in features to remain unchanged even when external conditions vary, especially clothing variations.
As shown in Fig. \ref{fig:intro} (b), it can be observed that when the model is able to focus on cloth-irrelevant regions, such as the head and legs, it can extract semantically consistent features that are more resistant to clothing changes.
Inspired by this, we propose a semantic-aware consistency network named \modelnameshort, which is a tri-stream mutual learning framework consisting of the raw image stream, head embedding stream, and clothing erasing stream.
\begin{comment}
Our basic idea is to simulate the visual attention mechanism of human so that the model can constantly focus on identity-related regions under clothing changes, thereby extracting more robust features.
\end{comment}
Firstly, to explicitly mitigate the interference of clothing variations, the human clothing from raw RGB images is removed to generate the black-clothing images.
The model learns color-based appearance features in the raw image stream, whereas in the clothing erasing stream, the model primarily learns cloth-independent features.
Secondly, in the head embedding stream, a head-enhanced attention module is designed, which utilizes the proposed part-based matching loss to learn soft attention maps from human body masks, highlighting the fine-grained identity-related head information.
Finally, to facilitate high-level common feature learning from three streams, a semantic consistency loss is presented to approximate the saliency maps of three streams, forcing the model to focus on semantically consistent identity-relevant regions.
The entire model is integrated into a unified framework for end-to-end training.
In the inference stage, only the features of the raw image stream are utilized, and the head embedding stream and clothing erasing stream are removed, reducing the reliance on the accuracy of human parsing and rendering our \modelnameshort computationally efficient and robust.

% The main contributions of this work are as follows:
The main contributions of this work are as follows:
\begin{itemize}
  \item We propose a novel tri-stream mutual learning framework called \modelnameshort for \ccreid task, which embeds appearance-based, identity-related, and cloth-irrelevant features into a unified network for joint optimization.
  \item To utilize fine-grained identity information, we design a head-enhanced attention module where the proposed part-based matching loss allows the model to learn soft attention maps more relevant to the \reid task. Besides, a semantic consistency loss is presented to constrain the model to focus on semantically invariant identity-relevant regions.
  \item Extensive experiments are conducted on four public \ccreid datasets, and the results show that the proposed \modelnameshort outperforms existing state-of-the-art methods.
\end{itemize}

\begin{figure*}[t]
  \centering
  \includegraphics[height=8.5cm,width=\linewidth]{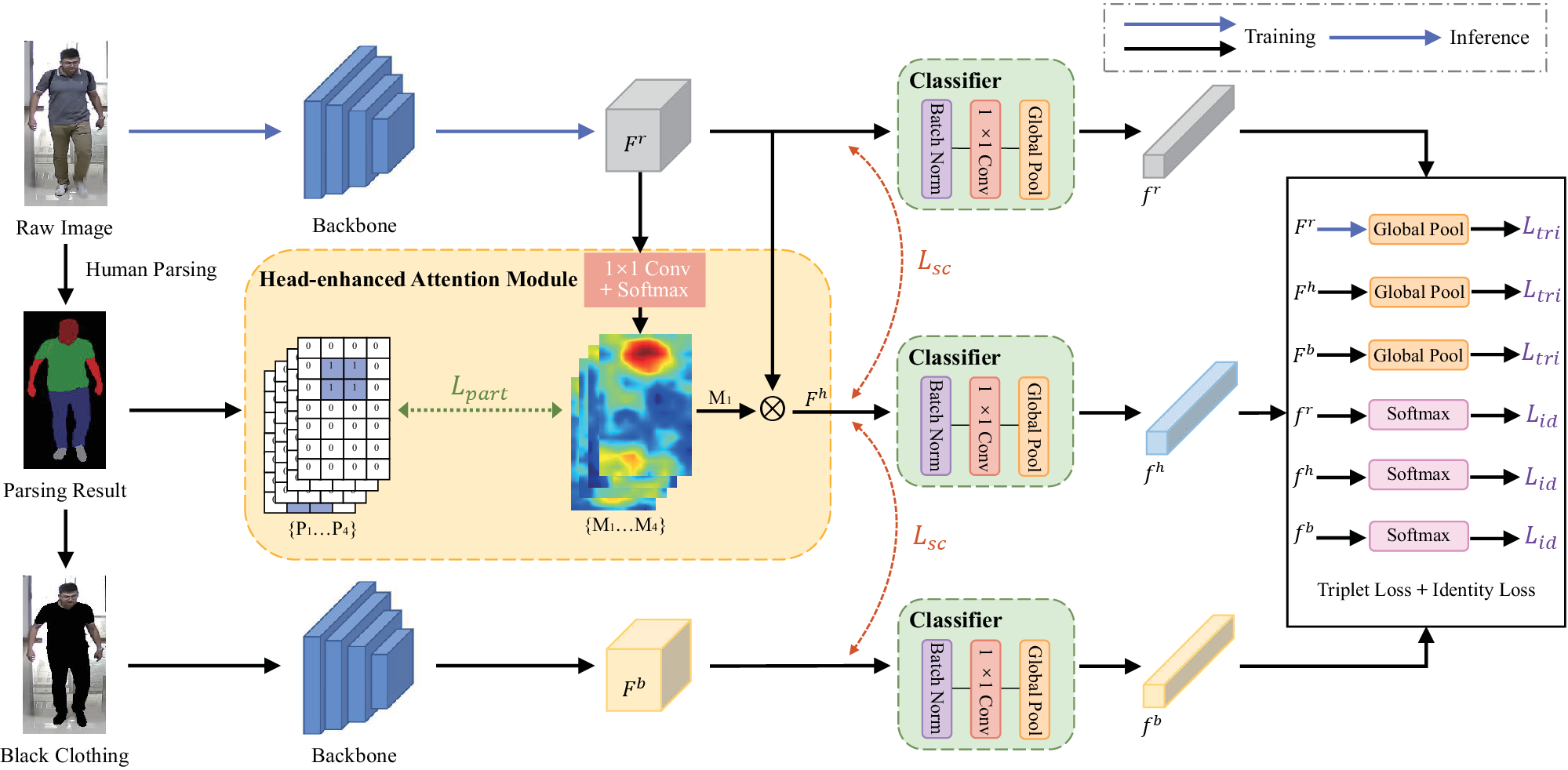}
  \caption{Overall architecture of the proposed tri-stream semantic-aware consistency network (\modelnameshort). From top to bottom are the raw image stream, the head embedding stream, and the clothing erasing stream, respectively. The raw RGB image and black-clothing image are fed into backbones that share weights, and the raw image stream features are input into the head-enhanced attention module that learns attention maps using the proposed part-based matching loss $\boldsymbol{L_{part}}$ to highlight the head information. The semantic consistency loss $\boldsymbol{L_{sc}}$ is presented to constrain the model to focus on semantically invariant identity-related regions. Meanwhile, the features in three streams are optimized by the triplet loss $\boldsymbol{L_{tri}}$ and identity loss $\boldsymbol{L_{id}}$.} 
  \label{fig:architecture}
\end{figure*}

\section{Related Work} \label{sec:2}
\begin{comment}
This section focuses on related works in the person \reid, including the general person \reid in Sec. \ref{sec:general} and the cloth-changing person \reid in Sec. \ref{sec:changing}.
\end{comment}

\subsection{General Person \reid} \label{sec:general}
Based on the assumption mentioned in Sec. \ref{sec:introduction}, almost all existing datasets, such as Market-1501 \cite{zheng2015scalable}, DukeMTMC \cite{ristani2016performance}, CUHK03 \cite{li2014deepreid}, and MSMT17 \cite{wei2018person} are captured in short time intervals without altering clothes for the same pedestrian.
Most current works mainly focus on the following types of problems in the general person \reid: pose changes \cite{ge2018fd, qian2018pose, su2017pose}, viewpoint differences \cite{jin2020uncertainty, sun2019dissecting}, illumination changes \cite{bak2018domain}, and occlusions \cite{huang2018adversarially, zhao2021incremental}. 
Song \etal \cite{song2018mask} propose a mask-guided contrast learning strategy and a region-level triplet loss for extracting discriminative features relevant to the identity.
An interaction-and-aggregation \reid framework \cite{hou2019interaction} is presented for modeling the interdependencies between spatial features and then aggregates the correlated features corresponding to the same body parts, enhancing the robustness to the variations in pose and scale.
Sun \etal \cite{sun2018beyond} propose a chunking-based feature extraction strategy, called PCB, which divides the pedestrian image equally into $p$ blocks in the vertical direction and computes classification loss on them separately to achieve fine-grained feature matching. 
In \cite{he2021transreid}, the authors propose a pure transformer network for the \reid task. 
They introduce a jigsaw patches module, which is used to reshuffle the patch embedding to improve the robustness and discriminative ability of the model.
These works have achieved excellent performance in the person \reid, as they heavily rely on the appearance information of human clothing that remains unchanged in the same pedestrian and provides an adequate basis for the model's judgment.
Nevertheless, as pedestrians change their clothes, the appearance information becomes unreliable, leading to a significant degradation in the performance of these methods.

\subsection{Cloth-changing Person \reid} \label{sec:changing}
As an increasing number of researchers become interested in \ccreid, some related datasets are released, such as LTCC \cite{qian2020long}, PRCC \cite{yang2019person}, VC-Clothes \cite{wan2020person}, Celebrities-ReID \cite{huang2019celebrities}, and NKUP \cite{wang2020benchmark}.
In these datasets, the same pedestrian changes two or more pieces of clothing and wears different accessories, such as the glass, hat, and backpack, which significantly enhances the appearance diversity and brings more challenges to this task.
To address the \ccreid, some works utilize auxiliary identity-related biological cues.
For example, Wang \etal \cite{wang2022co} extract appearance and body shape features from the original image and the heat map of body posture and then fuse them using the cross-attention mechanism.
Yang \etal \cite{yang2019person} release the PRCC dataset, which contains silhouette sketches of pedestrians as auxiliary discriminative information.
In \cite{jin2022cloth}, the authors present a consistency constraint to encourage a common feature learning from two modalities (dynamic gait and static RGB image).
Cui \etal \cite{cui2023dcr} disentangle clothes-relevant, clothes-irrelevant, and body contour features and help the model learn discriminative identity-related information by randomly shuffling the cloth-relevant embeddings.
In \cite{qian2020long}, the authors introduce a shape embedding module as well as a cloth-elimination shape-distillation module aiming to eliminate the now unreliable clothing appearance features and focus on the body shape information.
Some researchers also propose methods based on data augmentation \cite{shu2021semantic, xu2021adversarial, jia2022complementary} that enrich the color and texture of the clothes, facilitating the model to focus on cloth-irrelevant areas.
Additionally, several approaches based on metric learning \cite{li2022cloth, gu2022clothes} achieve significant improvements on \ccreid. 
Unlike existing works, our proposed method pays more attention to the learning of high-level identity-related semantic features and to constraining the model to focus on semantically consistent cloth-irrelevant regions, resulting in better robustness to clothing variations.

\section{METHODOLOGY} \label{sec:3}
\begin{comment}
This section presents the technical details of the proposed method. Firstly, we introduce the overall architecture of the semantic-aware consistency network in Sec.~\ref{sec:overall}. Then, we elaborate on the proposed head-enhanced attention module and semantic consistency loss in Sec. \ref{sec:head} and Sec. \ref{sec:sc}, respectively. Finally, we broadly describe the training and testing procedures of the model in Sec. \ref{sec:train}.
\end{comment}
\subsection{Overall Framework} \label{sec:overall}
As mentioned in Sec. \ref{sec:changing}, existing \ccreid methods utilize multi-modality features as auxiliary cues, and capturing and decoupling multi-modality information is usually time-consuming. 
Actually, the original RGB modality contains abundant cloth-irrelevant information that is underutilized by current approaches. 
To better mine the cloth-irrelevant information in the RGB modality, a tri-stream mutual learning framework is proposed.
As shown in Fig.~\ref{fig:architecture}, the model consists of three streams: the raw image stream, the head embedding stream, and the clothing erasing stream.
Since the information about clothes can interfere with the \ccreid, we use the pre-trained human body parsing network SCHPNet \cite{li2020self} and set the pixels corresponding to the clothes area in the parsing result to 0 to obtain images of pedestrians with black clothing.
The raw RGB image and the black-clothing image are fed into the backbone network of the raw image stream and clothing erasing stream to extract color-based appearance features $F^r$ and cloth-irrelevant features $F^b$, respectively.
The ResNet50 \cite{he2016deep} pre-trained on ImageNet \cite{krizhevsky2017imagenet} is used as the backbone network, and the backbones of both streams share weights.
Subsequently, the raw image stream features are fed into the head-enhanced attention module, which utilizes the presented part-based matching loss to encourage the learning of fine-grained head information.
After obtaining $F^r$ and $F^b$, and the head-enhanced features $F^h$, the proposed semantic consistency loss $L_{sc}$ is utilized to constrain saliency maps of all features to approximate each other, driving the model to focus on identity-related regions.
In addition, $F^r$, $F^h$, and $F^b$ are fed into a global average pooling layer to generate feature vectors that are utilized to compute the triplet loss \cite{hermans2017defense}.
Motivated by \cite{lin2013network}, we employ $1\times 1$ convolution followed by global average pooling instead of the standard fully connected layer.
A batch normalization layer is also added before the convolution layer to normalize features.
Since $1\times 1$ convolution with global average pooling possesses weaker classification capability, it allows the model to focus more on discriminative representation learning for accurate classification.
The identity loss $L_{id}$ \cite{sun2018beyond} is also utilized to supervise the embeddings $f^r$, $f^h$, and $f^b$ obtained after passing through the classifier, which can optimize the performance of identity identification.
In conclusion, the proposed \modelnameshort can guide the network to generate more consistent identity-related features that are well resistant to interference from clothing changes.

\subsection{Head-enhanced Attention Module} \label{sec:head}
The head-enhanced attention module takes the parsing results and features of the raw image stream $F^r$ as input and outputs the head-enhanced features $F^h$.
This module generates body part-based attention maps and human body masks, then learns the soft attention maps with the proposed part-based matching loss.
The architecture of this module is depicted in the center-left part of Fig. \ref{fig:architecture}.

\textbf{Body Part-based Attention Map.} In order to fully exploit fine-grained identity information, the $F^r\in \mathbb{R}^{H \times W \times C}$ is fed into the classifier composed of a $1 \times 1$ convolution layer followed by a softmax to generate attention maps $M \in \mathbb{R}^{H \times W \times K}$ , where $H$ denotes the height, $W$ denotes the width, and $C$ denotes the number of channels. 
The symbol $K$ denotes the number of attention maps, which is set to 4, corresponding to the head, upper body, lower body, and feet. 
The process can be formulated as:
\begin{equation}
M=\operatorname{softmax} (\operatorname{conv}(F^r)) ,
\end{equation}
where conv indicates the convolution operation. For the $k$-th attention map $M_k$, each pixel value represents the predicted probability that the current location belongs to the $k$-th body part, with values in the range $[0,1]$.

\textbf{Human Body Mask.} 
The human parsing result $Y \in \mathbb{R}^{H \times W}$ generated by SCHPNet~\cite{li2020self} has 18 labels that contain both cloth-relevant and cloth-irrelevant areas, such as upper clothes, pants, background, hair, face, and legs.
We change the pixel values of the head region in the parsing result to 1 and the other pixel values to 0, obtaining the mask of the head.
Other masks corresponding to the same areas as part-based attention maps are generated in the same way.
Ultimately, body part-based masks $P\in \mathbb{R}^{H \times W \times K}$ are utilized as supervision signals.

\textbf{Part-based Matching Loss.}
A new part-based matching loss $L_{part}$ is proposed to supervise the learning of attention maps from human body masks, and it can be defined as:
\begin{equation} \label{eq:part}
L_{part }=-\frac{1}{N} \frac{1}{H} \frac{1}{W} \sum_{n=1}^N \sum_{k=1}^K \sum_{h=0}^{H-1} \sum_{w=0}^{W-1} P_k(h, w) \log \left(M_k(h, w)\right) ,
\end{equation}
where $N$ is the batch size, $P_k(h, w)$ and $M_k(h, w)$ denotes the value of the $k$-th body mask and attention map at position $(h,w)$, respectively.
Notably, the learning process is also supervised by cross-entropy loss (identity loss), which employs identity labels.
Therefore, this module is trained under the part-based matching and person identification objectives.
Due to the supervision from identity loss, this module generates attention maps more relevant to the person \reid task than those generated using the fixed output of the pre-trained human parsing model.
Finally, since other body parts are inevitably affected by the clothing, we only perform element-wise multiplication of $M_1$ and $F^r$ in the spatial dimension to obtain the head-enhanced feature $F^h$.

\begin{figure}[t]
  \centering
  \includegraphics[width=\linewidth]{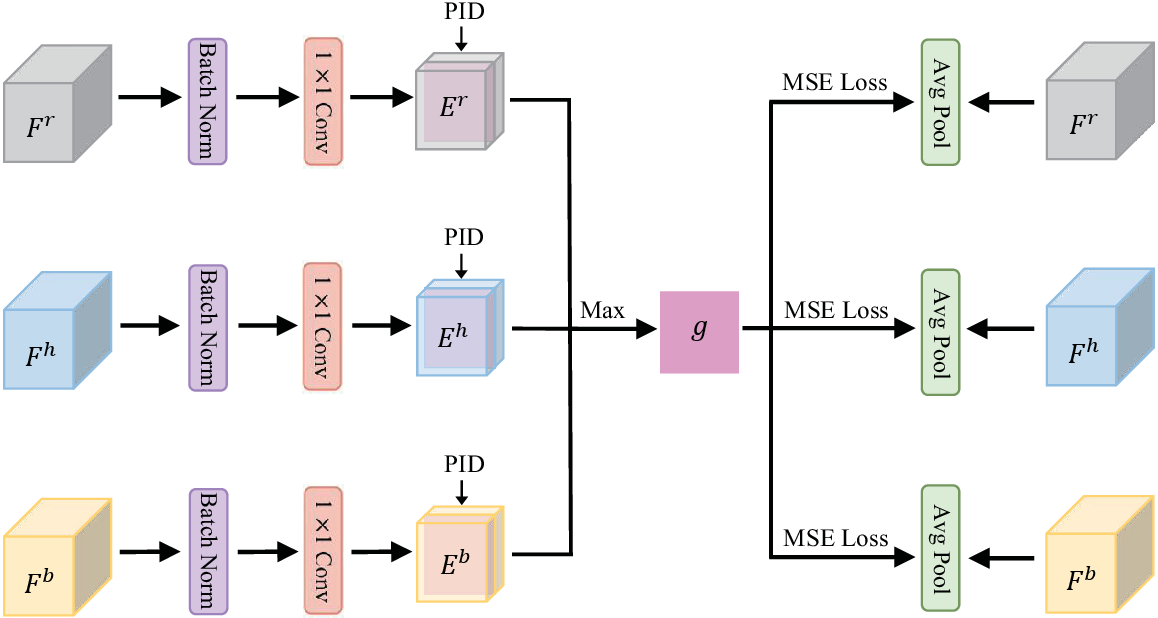}
  \caption{Illustration of the semantic consistency loss $\boldsymbol{L_{sc}}$. "PID" denotes the ground truth person ID.}
\label{loss:sc}
\end{figure}

\subsection{Semantic Consistency Loss}  \label{sec:sc}
In practical applications, the appearance of pedestrians usually changes, and only using traditional metric loss is not sufficient to help the model focus on semantically invariant regions without interference from clothing variations.
Therefore, a novel semantic consistency loss $L_{sc}$ is proposed to improve the robustness of the model to clothing changes, as illustrated in Fig. \ref{loss:sc}.
Specifically, $F^r$, $F^h$, and $F^b$ are fed into a batch normalization layer followed by a $1 \times 1$ convolution layer to obtain three class activation maps $E^r$, $E^h$, and $E^b \in \mathbb{R}^{H \times W \times I}$, where $I$ denotes the number of identity categories.
Considering the class activation map shows attention regions of the model, we select the feature map corresponding to the ground truth person ID $t$ in the channel dimension, thus obtaining $E_t^r$, $E_t^h$, and $E_t^b \in \mathbb{R}^{H \times W}$, respectively.
To further get a more effective supervision signal, we integrate the features from three streams, which can be expressed as:
\begin{equation}
g=\max(E_t^r, E_t^h, E_t^b) \xspace,
\end{equation}
where the $\max (\cdot)$ selects the maximum response value from three feature maps at the pixel level. 
Simultaneously, the average pooling on the features $F^r$, $F^h$, and $F^b$ along the channel dimension is performed separately to acquire saliency maps $F_a^r$, $F_a^h$, and $F_a^b \in \mathbb{R}^{H \times W}$, which essentially denote the focused regions by the network.
The semantic consistency loss $L_{sc}$ can be formulated as:
\begin{equation} \label{eq:sc}
L_{s c}=\frac{1}{N} \sum_{n=1}^N[(g-F_a^r)^2+(g-F_a^h)^2+(g-F_a^b)^2] \xspace.
\end{equation}

By approximating the saliency maps, the $L_{sc}$ facilitates the mutual learning between high-level semantic information from three streams.
This can help the raw image stream fully take advantage of the head-enhanced and cloth-irrelevant discriminative features from the other two streams, enhancing the distinguishing power under clothing variations.

\subsection{Training and Inference}  \label{sec:train}
Triplet loss \cite{hermans2017defense} and identity loss \cite{sun2018beyond} are commonly applied in the person \reid, where the triplet loss acts on the Euclidean space to improve the intra-class compactness and inter-class discriminability of features, and the identity loss acts on the cosine space to optimize the classification performance of the model.
In our experiments, besides $L_{part}$ and $L_{sc}$, both of the two losses $L_{id}$ and $L_{tri}$ are also adopted.
The overall training loss of the framework is defined as:
\begin{equation} \label{eq:total}
L=\lambda_1L_{part}+\lambda_2L_{s c}+\sum_{n \in Z}(L_{id}^n+L_{tri}^n) \xspace,
\end{equation}
where $Z=\{r, h, b\}$ corresponds to the raw, head-enhanced, and cloth-irrelevant features. Hyper-parameters $\lambda_1$ and $\lambda_2$ are used to balance the contribution of loss functions. 

The model is trained end-to-end without any additional training process. 
In the inference stage, the head embedding stream and the clothing erasing stream are discarded to save computational costs.

\begin{table*}[t]
\caption{Comparison of \modelnameshort with other state-of-the-art methods on LTCC, PRCC and Vc-Clothes datasets.}
\begin{center}
\setlength{\tabcolsep}{1.9mm}
\begin{tabular}{c|cccc|cccc|cccc}
\hline
\multirow{3}{*}{Methods} & \multicolumn{4}{c|}{LTCC} & \multicolumn{4}{c|}{PRCC} & \multicolumn{4}{c}{Vc-Clothes} \\ \cline{2-13} 
 & \multicolumn{2}{c|}{General} & \multicolumn{2}{c|}{Cloth-changing} & \multicolumn{2}{c|}{Same-clothes} & \multicolumn{2}{c|}{Cloth-changing} & \multicolumn{2}{c|}{General} & \multicolumn{2}{c}{Cloth-changing} \\ \cline{2-13} 
 & Rank-1 & \multicolumn{1}{c|}{mAP} & Rank-1 & mAP & Rank-1 & \multicolumn{1}{c|}{mAP} & Rank-1 & mAP & Rank-1 & \multicolumn{1}{c|}{mAP} & Rank-1 & mAP \\ \hline
HACNN (CVPR18)\cite{li2018harmonious} & 60.2 & \multicolumn{1}{c|}{26.7} & 21.6 & 9.3 & 82.5 & \multicolumn{1}{c|}{84.8} & 21.8 & 23.2 & 68.6 & \multicolumn{1}{c|}{69.7} & 49.6 & 50.1 \\
PCB (ECCV18)\cite{sun2018beyond} & 65.1 & \multicolumn{1}{c|}{30.6} & 23.5 & 10.0 & 99.8 & \multicolumn{1}{c|}{97.0} & 41.8 & 38.7 & 87.7 & \multicolumn{1}{c|}{74.6} & 62.0 & 62.2 \\
IANet (CVPR19)\cite{hou2019interaction} & 63.7 & \multicolumn{1}{c|}{31.0} & 25.0 & 12.6 & 99.4 & \multicolumn{1}{c|}{98.3} & 46.3 & 45.9 & - & \multicolumn{1}{c|}{-} & - & - \\
ISP (ECCV20)\cite{zhu2020identity} & 66.3 & \multicolumn{1}{c|}{29.6} & 27.8 & 11.9 & 92.8 & \multicolumn{1}{c|}{-} & 36.6 & - & 94.5 & \multicolumn{1}{c|}{94.7} & 72.0 & 72.1 \\ \hline
FSAM (CVPR21)\cite{hong2021fine} & 73.2 & \multicolumn{1}{c|}{35.4} & 38.5 & 16.2 & - & \multicolumn{1}{c|}{-} & - & - & 94.7 & \multicolumn{1}{c|}{\textbf{94.8}} & 78.6 & 78.9 \\
RCSANet (ICCV21)\cite{huang2021clothing} & - & \multicolumn{1}{c|}{-} & - & - & 100 & \multicolumn{1}{c|}{97.2} & 50.2 & 48.6 & - & \multicolumn{1}{c|}{-} & - & - \\
CAL (CVPR22)\cite{gu2022clothes} & 74.2 & \multicolumn{1}{c|}{40.8} & 40.1 & 18.0 & 100 & \multicolumn{1}{c|}{99.8} & 55.2 & 55.8 & 92.9 & \multicolumn{1}{c|}{87.2} & 81.4 & 81.7 \\
GI-ReID (CVPR22)\cite{jin2022cloth} & 63.2 & \multicolumn{1}{c|}{29.4} & 23.7 & 10.4 & - & \multicolumn{1}{c|}{-} & - & - & - & \multicolumn{1}{c|}{-} & 64.5 & 57.8 \\
M2Net (ACM MM22)\cite{liu2022long} & - & \multicolumn{1}{c|}{-} & - & - & 99.5 & \multicolumn{1}{c|}{99.1} & 59.3 & 57.7 & - & \multicolumn{1}{c|}{-} & - & - \\
AIM (CVPR23)\cite{yang2023good} & 76.3 & \multicolumn{1}{c|}{41.1} & 40.6 & 19.1 & 100 & \multicolumn{1}{c|}{\textbf{99.9}} & 57.9 & 58.3 & - & \multicolumn{1}{c|}{-} & - & - \\
CCFA (CVPR23)\cite{han2023clothing} & 75.8 & \multicolumn{1}{c|}{42.5} & 45.3 & 22.1 & 99.6 & \multicolumn{1}{c|}{98.7} & 61.2 & 58.4 & - & \multicolumn{1}{c|}{-} & - & - \\ \hline
\modelnameshort(Ours) & \textbf{76.3} & \multicolumn{1}{c|}{\textbf{43.6}} & \textbf{47.5} & \textbf{25.5} & \textbf{100} & \multicolumn{1}{c|}{97.8} & \textbf{61.3} & \textbf{59.9} & \textbf{94.9} & \multicolumn{1}{c|}{89.6} & \textbf{90.1} & \textbf{84.4} \\ \hline
\end{tabular}
\label{tab:sota}
\end{center}
\end{table*}

\begin{table}[]
\caption{Comparison of \modelnameshort with other state-of-the-art methods on the DeepChange dataset. This experiment is conducted under the general setting.}
\vspace{1mm}
\begin{center}
\setlength{\tabcolsep}{5.9mm}
\begin{tabular}{c|cc}
\hline
\multirow{2}{*}{Methods} & \multicolumn{2}{c}{DeepChange} \\ \cline{2-3} 
 & Rank-1 & mAP \\ \hline
MGN (ACM MM18)\cite{wang2018learning} & 25.4 & 9.8 \\
ABD-Net (ICCV19)\cite{chen2019abd} & 24.2 & 8.5 \\
OSNet (ICCV19)\cite{zhou2019omni} & 39.7 & 10.3 \\
RGA-SC (CVPR20)\cite{zhang2020relation} & 28.9 & 8.6 \\
ReIDCaps (TCSVT20)\cite{huang2019beyond} & 39.5 & 11.3 \\
Trans-reID (ICCV21)\cite{he2021transreid} & 35.9 & 14.4 \\
CAL (CVPR22)\cite{gu2022clothes} & \textbf{54.0} & \textbf{19.0} \\ \hline
\modelnameshort(Ours) & 53.5 & 18.7 \\ \hline
\end{tabular}
\label{tab:deepchange}
\end{center}
\end{table}

\section{Experiments} \label{sec:4}
\subsection{Datasets and Evaluation Metrics} \label{sec:data}
Four public cloth-changing datasets are selected to evaluate our approach, including LTCC \cite{qian2020long}, PRCC \cite{yang2019person}, VC-Clothes \cite{wan2020person}, and DeepChange \cite{xu2021deepchange}. 
The LTCC dataset is an indoor cloth-changing dataset, which has 17,138 images of 152 identities with 478 different outfits. 
The multiplied scenarios (collected by 12 cameras), variants of clothes (up to 14 changes of clothes for each person), and significant illumination variations make it one of the most challenging \ccreid datasets in recent years. 
The PRCC dataset is the other cloth-changing dataset containing 33,698 images from 221 persons, taken by 3 cameras. 
Each person in cameras A and B wears the same clothes, and the person under camera C wears different clothes from A and B. 
The VC-Clothes dataset is a synthetic cloth-changing dataset rendered by the GTA5 game engine, which contains 19,060 images of 512 identities captured from 4 cameras. 
Each identity has $1\sim3$ suits of clothes. 
The DeepChange dataset is a large-scale cloth-changing dataset under real scenarios, which has 12 months of surveillance data covering clothing and behavioral changes. 
It is captured by 17 surveillance cameras of different resolutions and contains 171,352 images of 1,082 pedestrian identities.

Rank-1 accuracy and mean average precision (mAP) are used as evaluation metrics. 
Following \cite{gu2022clothes}, the three test settings are defined as (1) \textbf{general setting:} both clothes-changing and clothes-consistent gallery samples are used to calculate accuracy, (2) \textbf{cloth-changing setting:} only clothes-changing gallery samples are used to calculate accuracy, and (3) \textbf{same-clothes setting:} only clothes-consistent gallery samples are used to calculate accuracy. 
For LTCC and Vc-Clothes, we report the Rank-1 accuracy and mAP under the general and clothes-changing settings.
As for PRCC, the Rank-1 accuracy and mAP under the same-clothes and clothes-changing settings are reported \cite{gu2022clothes, yang2019person}. 
For DeepChange, we report the Rank-1 accuracy and mAP under the general setting \cite{gu2022clothes}.

\subsection{Implementation Details}  \label{sec:detail}

The ResNet50 \cite{he2016deep} pre-trained on ImageNet \cite{krizhevsky2017imagenet} is adopted as the backbone for the raw image stream and clothing erasing stream, and we remove the global average pooling layer and the final fully connected layer. 
Following the general routine \cite{luo2019bag}, the stride of the last stage in ResNet50 is set to 1 instead of 2, which improves the spatial resolution of extracted features. 
The input images are resized to $384\times192$ following \cite{qian2020long}. 
Random horizontal flipping, random cropping, and random erasing \cite{zhong2020random} are used for data augmentation. 
The training batch size is 32, with 4 pedestrians and 8 images for each pedestrian.
The model is trained for 150 epochs using the Adam optimizer. 
We use the warm-up learning rate strategy \cite{luo2019bag}, $\ie$, the learning rate increases linearly from $3.5\times 10 ^{-6}$ to $3.5\times 10 ^{-4}$ in the first 10 epochs, and it is divided by 10 in the 40th epoch and 80th epoch, respectively.
$\lambda_1$ and $\lambda_2$ in Eq. \eqref{eq:total} are both set to 0.01 on LTCC, Vc-Clothes, and DeepChange datasets and to 0.1 on the PRCC dataset. 
The triplet loss margin $\alpha$ is set to 0.3. 

\subsection{Comparison with State-of-the-art Methods} \label{sec:compare}
We compare our \modelnameshort with state-of-the-art methods on LTCC, PRCC, and Vc-Clothes datasets. 
The methods used for comparison include four general \reid methods ($\ie$ HACNN \cite{li2018harmonious}, PCB \cite{sun2018beyond}, IANet \cite{hou2019interaction}, and ISP \cite{zhu2020identity}) and seven \ccreid methods ($\ie$ FSAM \cite{hong2021fine}, RCSANet \cite{huang2021clothing}, CAL \cite{gu2022clothes}, GI-ReID \cite{jin2022cloth}, M2Net \cite{liu2022long}, AIM \cite{yang2023good} and CCFA \cite{han2023clothing}). 
The results are reported in Table \ref{tab:sota}.
It can be seen that most of \ccreid approaches achieve better performance than general \reid approaches due to the addition of auxiliary modules or optimization functions to resist clothing variations. 
Compared with these methods, our proposed \modelnameshort can make sure that the model focuses on cloth-irrelevant areas through effective semantic consistency loss, obtaining features more robust to appearance changes. 
As a result, our method achieves significant performance improvements over SOTA methods under the cloth-change setting. 
Specifically, on the LTCC dataset, our method outperforms sub-optimal CCFA by 2.2\% and 3.4\% on Rank-1 and mAP, respectively.
On the PRCC dataset, the proposed \modelnameshort achieves better performance than the SOTA method CCFA.
Our method also outperforms CAL by 8.7\% Rank-1 and 2.7\% mAP on Vc-Clothes.
Although our method is specific to the \ccreid, our method still obtains competitive results under the general setting, indicating that the proposed \modelnameshort is able to fully explore the fine-grained identity-related information without the interference of clothes. 
Besides, our proposed \modelnameshort only uses features of the raw image stream in the inference stage without using human parsing for computational efficiency.
We also compare the proposed \modelnameshort with the state-of-the-art methods on the DeepChange dataset, and the results are shown in Table \ref{tab:deepchange}. 
The best performance is obtained by CAL, with 54.0\% Rank-1 and 19.0\% mAP, and our approach also achieves competitive results. 
The reason why our \modelnameshort is not superior to CAL on DeepChange is that the faces of pedestrian images in the DeepChange dataset are blurred for privacy, which largely weakens the discrimination of the head region, resulting in performance degradation of the proposed method.

\subsection{Ablation Study} \label{sec:ablation} 
In this section, ablation studies are performed to verify the effectiveness of our proposed method.
Firstly, we investigate the effect of each stream. 
Secondly, we study the impact of loss functions on model performance. 
Then, the influence of using different classifiers is explored. 
Finally, we show the effect of hyper-parameters. 
Except for the experiments of hyper-parameters, all experiments are conducted under the cloth-changing setting.

\begin{table}[t]
\caption{Ablation experiments of streams on LTCC and PRCC datasets. "Raw", "Head" and "Black" indicate the raw image stream, the head embedding stream, and the clothing erasing stream, respectively.}
\setlength{\tabcolsep}{3.3pt}
\vspace{1.7mm}
\begin{tabular}{c|ccc|cc|cc}
\hline
\multirow{2}{*}{Methods} & \multirow{2}{*}{Raw} & \multirow{2}{*}{Head} & \multirow{2}{*}{Black} & \multicolumn{2}{c|}{LTCC} & \multicolumn{2}{c}{PRCC} \\ \cline{5-8} 
 &  &  &  & Rank-1 & mAP & Rank-1 & mAP \\ \hline
1 (Baseline) & \checkmark & $\times$ & $\times$ & 33.8 & 15.0 & 46.3 & 43.2 \\
2 & $\times$ & \checkmark & $\times$ & 34.2 & 15.6 & 46.8 & 43.4 \\
3 & $\times$ & $\times$ & \checkmark & 34.6 & 16.1 & 47.3 & 44.8 \\
4 & \checkmark & \checkmark & $\times$ & 35.3 & 16.6 & 47.0 & 43.9 \\
5 & \checkmark & $\times$ & \checkmark & 35.1 & 16.3 & 47.9 & 45.4 \\ 
6 & $\times$ & \checkmark & \checkmark & 37.4 & 17.5 & 48.3 & 46.9 \\
7 & \checkmark & \checkmark & \checkmark & \textbf{40.2} & \textbf{20.5} & \textbf{52.6} & \textbf{51.7} \\ \hline
\end{tabular}
\label{tab:modal}
\end{table}

\textbf{Effect of streams.}
As shown in Table \ref{tab:modal}, compared with method 1 ($\ie$ baseline, which uses the original image as the input of ResNet50 and utilizes identity loss for training), method 2 can extract fine-grained identity-related features in the head region, so it achieves better performance on both datasets.  
Method 3 has more significant performance improvements due to the complete masking of clothing-related regions. 
The results from methods 4,5,6 show that better performance can be achieved using fused features in the dual-stream model.
It is observed that method 3 gets better performance than method 4 on the PRCC dataset.
This is likely due to the fact that the PRCC dataset contains smaller occlusions, viewpoint differences, and illumination variations and that clothing changes are the dominant factor affecting appearance characteristics, so removing human clothing can considerably improve performance.
Finally, fusion features of three streams in \modelnameshort achieve the best performance ($\ie$ method 7), outperforming the baseline 6.4\%/6.3\% Rank-1 and 5.5\%/8.5\% mAP on both datasets. 
It illustrates the benefit of multi-stream information for person \reid. 
Notably, in this ablation experiment, the head-enhanced features in the head embedding stream are not obtained with the proposed part-based matching loss but directly using attention maps provided by the human parsing results.
In addition, only identity loss is used to optimize all networks.
In the training and inference stages, the fusion of features in different streams is performed by concatenating them in the channel dimension.
Due to the direct concatenation of features and the lack of effective constraints, the model does not focus only on cloth-irrelevant regions.
In the following experiment, the impact of the proposed constraints is investigated.

\begin{table}[t]
\caption{Ablation experiments of loss functions on LTCC and PRCC datasets. "Tri", "Pm" and "Sc" denote the triplet loss, part-based matching loss (Eq. \eqref{eq:part}), and semantic consistency loss (Eq. \eqref{eq:sc}), respectively.}
\setlength{\tabcolsep}{5.5pt}
\begin{tabular}{c|ccc|cc|cc}
\hline
\multirow{2}{*}{Methods} & \multirow{2}{*}{Tri} & \multirow{2}{*}{Pm} & \multirow{2}{*}{Sc} & \multicolumn{2}{c|}{LTCC} & \multicolumn{2}{c}{PRCC} \\ \cline{5-8} 
 &  &  &  & Rank-1 & mAP & Rank-1 & mAP \\ \hline
1 & $\times$ & $\times$ & $\times$ & 40.2 & 20.5 & 52.6 & 51.7 \\
2 & \checkmark & $\times$ & $\times$ & 40.8 & 21.2 & 53.3 & 52.0 \\
3 & \checkmark & \checkmark & $\times$ & 42.6 & 23.2 & 56.7 & 54.8 \\
4 & \checkmark & $\times$ & \checkmark & 44.9 & 24.1 & 58.5 & 56.0 \\
5 & \checkmark & \checkmark & \checkmark & \textbf{47.5} & \textbf{25.5} & \textbf{61.3} & \textbf{59.9} \\ \hline
\end{tabular}
\label{tab:loss}
\end{table}

\begin{table}[t]
\caption{Ablation experiments of classifiers on LTCC and PRCC datasets.}
\setlength{\tabcolsep}{5.9pt}
\begin{tabular}{c|c|cc|cc}
\hline
\multirow{2}{*}{Methods} & \multirow{2}{*}{Classifiers} & \multicolumn{2}{c|}{LTCC} & \multicolumn{2}{c}{PRCC} \\ \cline{3-6} 
 &  & Rank-1 & mAP & Rank-1 & mAP \\ \hline
\multirow{2}{*}{1 (Baseline)} & BNNeck & \textbf{34.3} & \textbf{15.7} & \textbf{46.9} & 43.0 \\
 & BGAP & 33.8 & 15.0 & 46.3 & \textbf{43.2} \\ \hline
\multirow{2}{*}{2 (\modelnameshort)} & BNNeck & 46.3 & 24.7 & 60.2 & 59.0  \\
 & BGAP & \textbf{47.5} & \textbf{25.5} & \textbf{61.3} & \textbf{59.9} \\ \hline
\end{tabular}
\label{tab:cla}
\end{table}

\textbf{Effect of loss functions.} We explore the effect of different loss functions on model performance, and the results are shown in Table \ref{tab:loss}, where all methods utilize features from three streams.
Compared with method 1 which only uses identity loss, method 2 adds triplet loss to optimize the model on the Euclidean space, enhancing the intra-class compactness and inter-class discriminability of features. 
Method 3 outperforms 1.8\%/3.4\% Rank-1 and 2.0\%/2.8\% mAP than method 2 on LTCC and PRCC, respectively, which demonstrates that under the supervision of part matching and identity recognition, the head-enhanced attention module can generate soft attention maps that are more relevant to the Re-ID task, selectively preserving identity-related features.  
According to the results of method 2 and method 4, we observe that the proposed semantic consistency loss $L_{sc}$ improves the 4.1\%/5.2\% Rank-1 and 2.9\%/4.0\% mAP on both datasets, which shows that $L_{sc}$ can consistently constrain the model to focus on identity-related regions, better resisting the interference of clothing variations. 
Finally, the best performance is achieved by simultaneously using identity loss, triplet loss, part-based matching loss, and semantic consistency loss ($\ie$ method 5), illustrating the effectiveness of the proposed method for \ccreid.

\begin{figure}[t]
  \centering
  \includegraphics[width=\linewidth, height=3.5in]{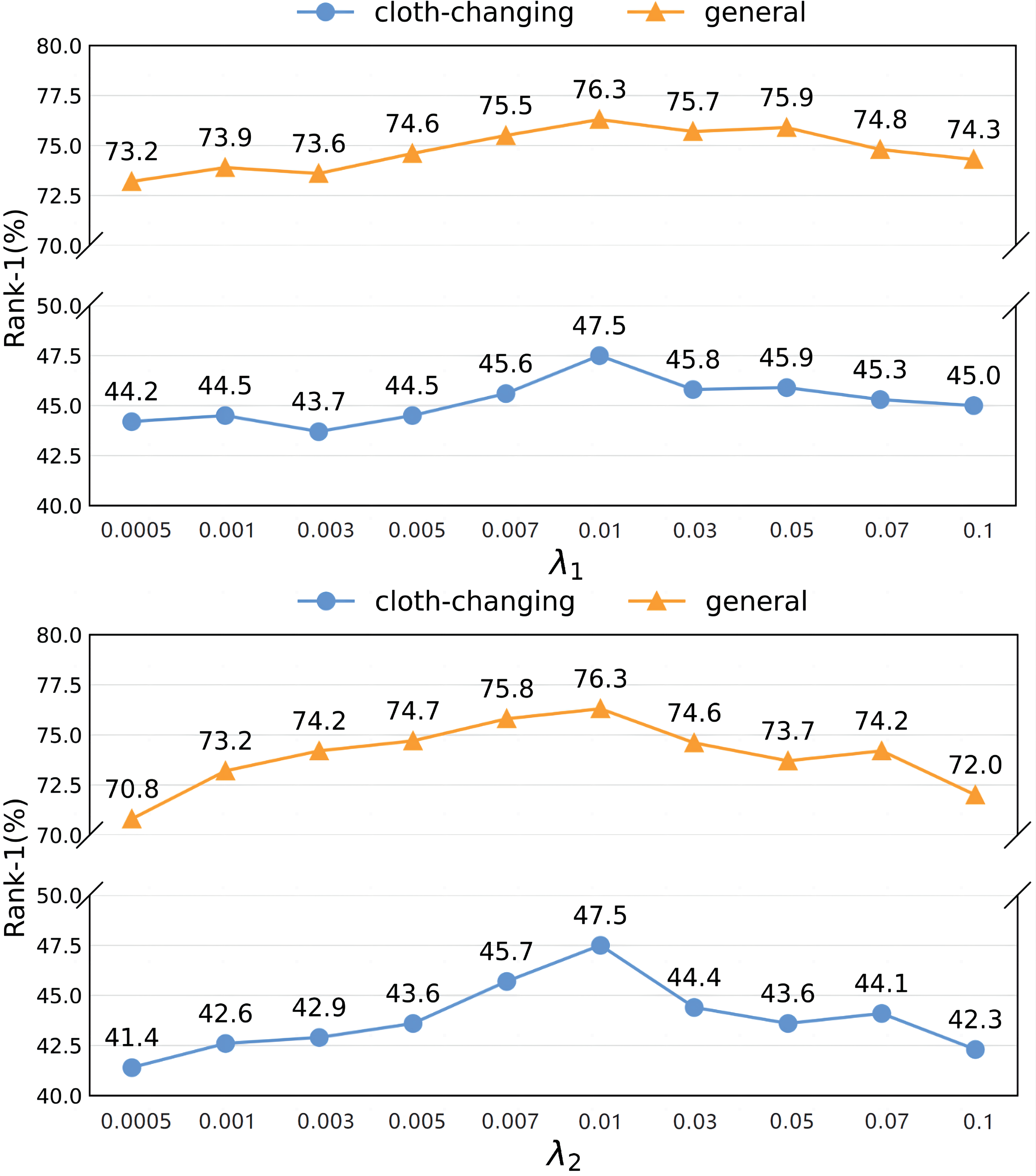}
  \caption{Ablation experiments of hyper-parameters on the LTCC dataset. Above: the impact of $\boldsymbol{\lambda_1}$ on Rank-1 accuracy. Below: the impact of $\boldsymbol{\lambda_2}$ on Rank-1 accuracy.}
  \label{fig:lambda}
\end{figure}

\textbf{Effect of classifiers.}
BNNeck \cite{luo2019bag} is a widely used classifier in the person \reid, which consists of a batch normalization layer and a fully connected layer.
It uses triplet loss before the batch normalization layer and id loss after the fully connected layer, thus optimizing the features in both Euclidean space and cosine space simultaneously, significantly improving the performance of the \reid model.
Instead of using BNNeck, following \cite{lin2013network}, we use the $1\times1$ convolution with global average pooling to replace the fully connected layer. 
Besides, a batch normalization layer is added before the $1\times1$ convolution layer to facilitate model convergence. 
We abbreviate the classifier composed of these layers as BGAP.
The comparison results of BNNeck and BGAP are shown in Table \ref{tab:cla}.
It is observed that for baseline, the performance of BNNeck is generally better.
But for our proposed \modelnameshort, using BGAP instead of BNNeck can achieve superior results.
It might be because BGAP has weaker discriminative power than BNNeck and requires more powerful features for accurate classification.
This encourages our \modelnameshort to focus on feature learning during end-to-end training, resulting in improved performance.

\textbf{Effect of hyper-parameters.} There are two hyper-parameters $\lambda_1$ and $\lambda_2$ for balancing the weights of the loss function, as shown in Eq. \eqref{eq:total}. 
To explore the impact of these hyper-parameters on the model performance, we conduct experiments on the LTCC dataset under the general setting and cloth-changing setting. 
When evaluating one of the parameters, the other hyper-parameter is fixed as the optimal value. 
The results are shown in Fig. \ref{fig:lambda}. 
The model achieves the best performance when both $\lambda_1$ and $\lambda_2$ are equal to 0.01.
It can also be observed that the model accuracy generally shows an increasing trend with $\lambda_1$ or $\lambda_2$ becoming larger, while it begins to decrease when $\lambda_1$/$\lambda_2$ is greater than 0.01. 
Therefore, $\lambda_1=\lambda_2=0.01$ is a reasonable choice. 

\begin{figure}[t]
  \centering
  \includegraphics[width=\linewidth, height=3.66in]{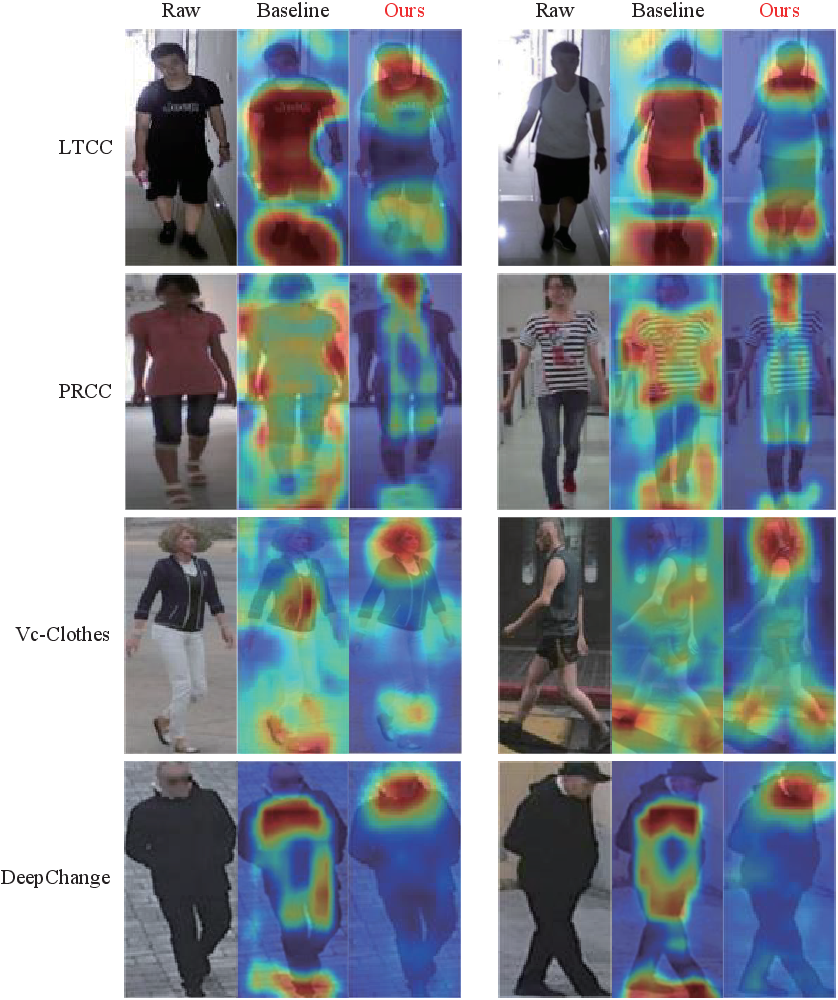}
  \caption{Visualization of raw person images and activation maps learned by the baseline and our \modelnameshort.}
  \label{fig:attention}
\end{figure}

\subsection{Visualization and Analysis} \label{sec:visual}
\textbf{Visualization of activation maps.} To verify which cues the model has learned, we visualize the activation maps of baseline and proposed \modelnameshort for comparison, as shown in Fig. \ref{fig:attention}.
Each row from top to bottom is the visualization result on LTCC, PRCC, Vc-Clothes, and DeepChange, respectively.
The activation maps can indicate which characteristics the model pays more attention to.
From the results, it can be seen that the highlights of activation maps learned by baseline are scattered, whereas the highlights of activation maps learned by \modelnameshort are more centralized, which demonstrates that the \modelnameshort is subject to more constraints in the process of representation learning.
It can also be observed that the baseline focuses more on cloth-relevant regions such as the clothes and shoes, while \modelnameshort concentrates more on cloth-irrelevant regions such as the head and legs.
This validates the effectiveness of the proposed head-enhanced attention module and semantic consistency loss for the \ccreid task. 

As shown in the visualization results on four datasets, our proposed method is able to accurately focus on identity-related regions in different scenarios, camera views, poses, and clothes.
However, the baseline is easily affected in these situations.
It shows that the \modelnameshort is able to extract features that are more robust to the above changes.
Moreover, despite the lower resolution of PRCC and DeepChange datasets, our approach is still capable of precisely focusing on the head region, which demonstrates the feasibility of the proposed \modelnameshort in intelligent surveillance applications.

\begin{figure}[t]
  \centering
  \includegraphics[width=\linewidth, height=3.2in]{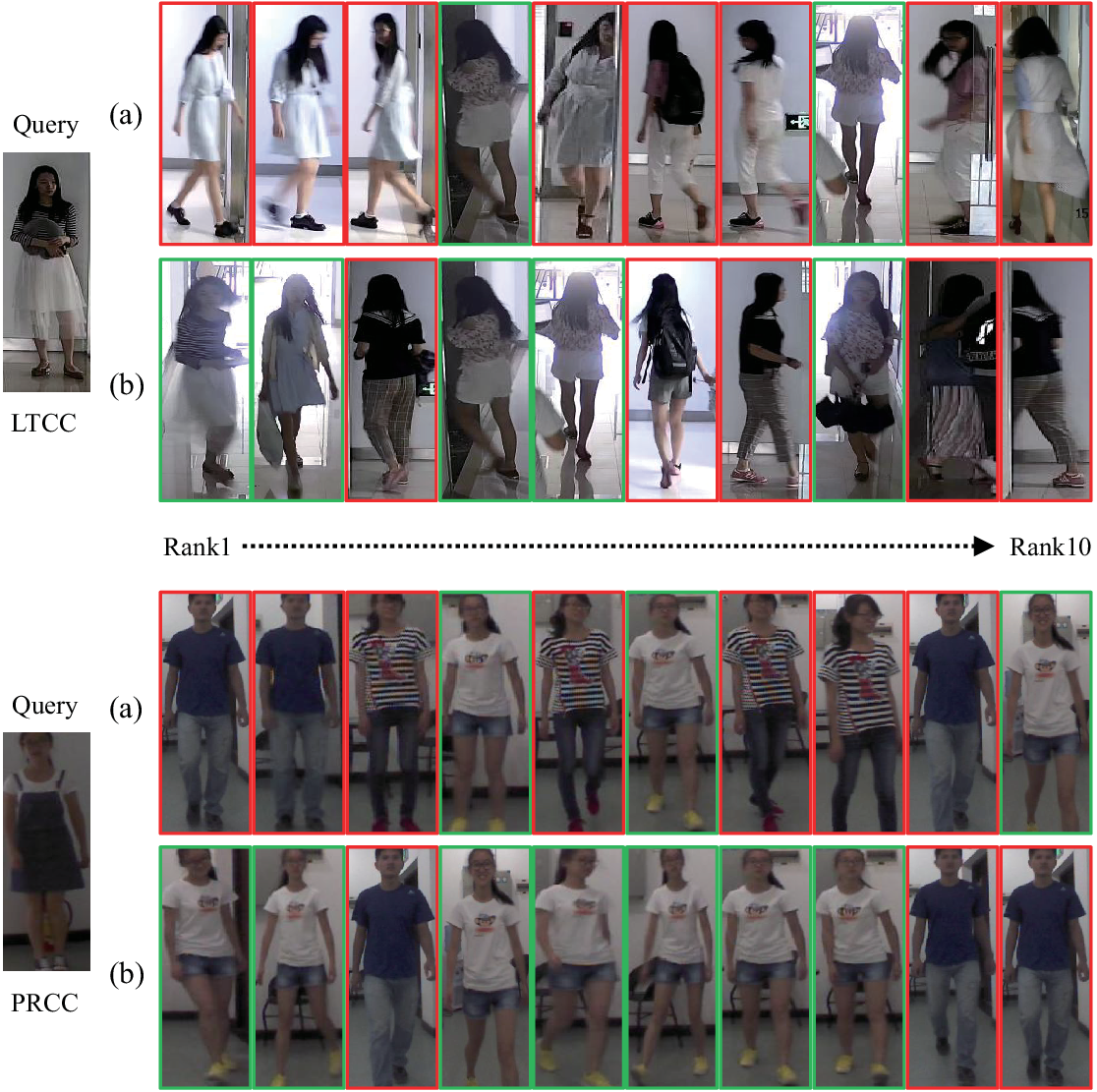}
  \caption{Visualization of the top-10 ranking lists by baseline and our \modelnameshort on the LTCC and PRCC datasets. Images in green and red boxes are positive and negative results, respectively. (a) Retrieval results produced by baseline. (b) Retrieval results produced by \modelnameshort.}
  \label{fig:ranking}
\end{figure}

\textbf{Visualization of retrieval results.} To give an intuitive evaluation of our proposed \modelnameshort, we visualize the top-10 ranked retrieval results of the baseline network and our \modelnameshort under the cloth-changing setting on LTCC and PRCC datasets, as shown in Fig. \ref{fig:ranking}.
The experiment is based on the cloth-changing setting, so gallery samples that have the same identity and the same clothes as the query are removed.
It can be observed that the LTCC dataset has more dramatic clothing changes, illumination variations, and pose changes compared with the PRCC dataset, and the results show that our method exhibits better robustness than the baseline model on both datasets.
For example, the ranking lists in the first row indicate that the baseline is disturbed by the color of the clothes and mismatches the person wearing white dresses or white pants.
The ranking lists in the third row also show that the baseline is affected by the texture of the clothes and incorrectly matches pedestrians wearing clothes with stripes.
However, benefiting from the proposed head-enhanced attention module and semantic consistency loss, our approach effectively resists interferences from the color and texture of clothes, showing great retrieval quality. 

\section{Conclusion} \label{sec:5}
In this paper, we suggest utilizing the effective consistency constraint to address the \ccreid task.
A novel tri-stream framework named \modelnameshort is proposed, which exploits cloth-irrelevant features and head-enhanced features (discarded in the inference) to encourage identity-related representation learning.
To extract cloth-irrelevant features, we generate black-clothing images by fully erasing human clothing.
We also propose a head-enhanced attention module to exploit fine-grained identity information.
A semantic consistency loss is further designed to facilitate the learning of semantically consistent identity-related features, which significantly improves the robustness of the model to clothing variations.
Extensive experiments on four public \ccreid datasets show that the proposed \modelnameshort achieves state-of-the-art performance.
The limitation of our method is that it partially relies on the accuracy of human parsing results, which will be improved in future work.

\begin{acks}
    This work is supported by the National Natural Science Foundation of China (No.62073004), Shenzhen Fundamental Research Program (No.GXWD20201231165807007-20200807164903001) and Science and Technology Plan of Shenzhen (No.JCYJ20200109140410340).
\end{acks}

% \begin{acks}
%To Robert, for the bagels and explaining CMYK and color spaces.
% \end{acks}

\bibliographystyle{ACM-Reference-Format}
\balance
\bibliography{ref}

\appendix
\end{sloppypar}
\end{document}